# Polarimetric Hierarchical Semantic Model and Scattering Mechanism Based PolSAR Image Classification

Fang Liu[a,b,c], Junfei Shi[a,b,c], Licheng Jiao[b,c], Hongying Liu[b,c],

Shuyuan Yang[b,c], Jie Wu[a,b,c], Hongxia Hao[a,b,c], Jialing Yuan[a,b,c]

[a]*School of Computer Science and Technology, Xidian University, Xi'an, 710071, P.R. China*

[b]*Key Laboratory of Intelligent Perception and Image Understanding of Ministry of Education of China, Xidian University, Xi'an 710071, P.R. China*

[c]*International Center of Intelligent Perception and Computation, Xidian University, Xi'an, 710071, P.R. China*

**Abstract:** For polarimetric SAR (PolSAR) image classification, it is a challenge to classify the aggregated terrain types, such as the urban area, into semantic homogenous regions due to sharp bright-dark variations in intensity. The aggregated terrain type is formulated by the similar ground objects aggregated together. In this paper, a polarimetric hierarchical semantic model (PHSM) is firstly proposed to overcome this disadvantage based on the constructions of a primal-level and a middle-level semantic. The primal-level semantic is a polarimetric sketch map which consists of sketch segments as the sparse representation of a PolSAR image. The middle-level semantic is a region map which can extract semantic homogenous regions from the sketch map by exploiting the topological structure of sketch segments. Mapping the region map to the PolSAR image, a complex PolSAR scene is partitioned into aggregated, structural and homogenous pixel-level subspaces with the characteristics of relatively coherent terrain types in each subspace. Then, according to the characteristics of three subspaces above, three specific methods are adopted, and furthermore polarimetric information is exploited to improve the segmentation result. Experimental results on PolSAR data sets with different bands and sensors demonstrate that the proposed method is superior to the state-of-the-art methods in region homogeneity and edge preservation for terrain classification.

# I INTRODUCTION

Terrain classification is one of the main tasks for polarimetric synthetic aperture radar (PolSAR) image processing, and it is the base of the subsequent image understanding and interpretation. With the rapid accumulation of PolSAR data, how to effectively and intelligently classify the complex scene of PolSAR images with hybrid terrain types has become an important and common concern. Moreover, due to the lack of the labeled data in PolSAR images, the unsupervised terrain classification attracts more attentions than the supervised terrain classification.

A number of unsupervised classification approaches [1-11] are popular by making full use of polarimetric information and mainly categorized into three types: 1) scattering mechanism based methods [2-4] which is mainly based on target decomposition theory; 2) statistical distribution based approaches [5-8] which assume the PolSAR data as the Gaussian or non-Gaussian distributions; 3) methods [9-11] by combining the scattering characteristics and statistical models, such as the H/$\alpha$-Wishart method [9]. These approaches take advantages of scattering characteristics and statistical distribution of PolSAR data, and can obtain fine classification results without prior knowledge or labels. However, without considering pixels' spatial relationship, these pixel-based classification methods usually produce an inconsistent salt-and-pepper classification result.

To take the spatial information into account, the image segmentation techniques are involved for PolSAR classification in recent years. There are plenty of segmentation methods to exploit spatial information which can be mainly divided into four categories: 1) approaches based on superpixels which are over-segmented regions, such as the hierarchical segmentation [12] and region-based methods [13-15]; 2) approaches based on textural modeling[16], such as gray level co-occurrences matrices (GLCM) [17] and Gabor [18] or wavelet features [19]; 3) approaches with regularization criterion, such as Markov Random Field(MRF) [20][21] and contour criterion [22][23]; 4) approaches based on statistical modeling, such as the non-Gaussian modeling method [24]. Compared with classification methods without spatial information, these approaches which consider the local dependency in an image can achieve more homogenous classification results. However, these spatial information-based methods are difficult to produce both semantic homogenous regions and accurate edges especially for the aggregated terrain types due to the sharp bright-dark variations in intensity.

The aggregated terrain types, such as the urban area or the forest, are formulated by similar ground objects aggregated together. Their major characteristic is the bright-dark variations in intensity which are caused by scattering waves of objects and the close ground in low-resolution images. For instance, Fig. 1(a) is the total backscattering power (SPAN) image of San Francisco area, and Fig. 1(c) is the subimage of the urban area from Fig. 1(a). As shown in Fig. 1(c), the building and its surrounding ground can formulate sharp bright-dark variation in intensity. Moreover, the bright-dark variations in intensity can be also caused by scattering waves of objects and their shadows in high-resolution SAR images, such as the forest in the Fig. 2(c). For the low-resolution PolSAR image such as Fig. 1(c), since the buildings are clustered together, the bright-dark variations appear repeatedly. Various merging schemes are difficult to segment the buildings and its surrounding ground into homogenous regions with local features. However, to better understand the PolSAR scene, the urban area should be segmented into a whole region from human's interpretation. It is a contradiction between image's low-level features and the high-level semantic which can characterize the image object in the view of vision. Therefore, to solve this contradiction and obtain semantic homogenous aggregated regions, high-level semantic features should be exploited to better represent the aggregated ground objects.

The visual perception of humans [25] has the capability of hierarchical cognition [26] which can recognize the aggregated terrain as a homogenous region. Learning from the sparse cognition characteristics of humans' vision, some high level features [27-29] and semantic information [30][31] should be exploited for POLSAR images. Primal sketch model is such a tool to obtain the sparse representation of the image. Based on Marr's vision theory [32], a primal sketch model [27] was implemented to obtain a sketch map as the sparse representation of the image. In the sketch map, sketch segments are formulated due to the bright-dark variation in intensity. According to the sketch map, an image can be partitioned into sketchable and non-sketchable part**s.** However, the primal sketch model was proposed for natural images whose intensity changes always appear in edges. With a dictionary of SAR image primitives, Wu and Liu [33] developed the primal sketch model for the SAR image to reduce speckle and sound performance is obtained.

Inspired by the primal sketch model, we propose a polarimetric hierarchical semantic model (PHSM) to partition a PolSAR image into aggregated, structural and homogenous regions. The PHSM consists of two levels: the primal-level semantic is a polarimetric sketch map and the middle-level semantic is a region map. Specifically, considering polarimetric characteristics, a polarimetric sketch map is exploited as a sparse representation of PolSAR images. Fig. 1(b) and Fig. 2(b) give schematic illustrations of polarimetric sketch maps. As shown in Fig. 1(b) and Fig. 2(b), the sketch segments can not only represent the edges and line objects of PolSAR images, but also describe the bright-dark variation in the aggregated terrain type well. Therefore, the sketch segments can be used to describe the aggregated terrain types. In addition, we further extract the region map on the polarimetric sketch map by exploiting the spatial relationship of sketch segments. The region map partitions a PolSAR image into aggregated, structural and homogenous regions, and aggregated regions guarantee the aggregated terrain types as semantic homogenous regions.

Moreover, different models can be designed for three regions by considering their own characteristics.

In this paper, based on the PHSM, an unsupervised terrain classification approach which combines polarimetric hierarchical semantic with the scattering mechanism (PHS_SM) is proposed for the POLSAR imagery. The PHS_SM method shows different characteristics from the conventional classification methods as follows. 1) It exploits hierarchical semantic models to divide a PolSAR image into aggregated, structural and homogenous subspaces to obtain different semantic homogenous regions. 2) Considering the different characteristics of the three subspaces, three different segmentation schemes are designed. 3) For the segmentation map to get accurate label, a semantic-polarimetric classifier is adopted which combines the semantic information-based segmentation with polarimetric mechanism-based classification. The PHS_SM method can obtain better performance over other state-of-art methods on both region homogeneity and edge preservation.

The outline of this paper is organized as follows. In Section II, the background and motivation are introduced respectively. In Section III, the proposed PHSM is described in detail. The PHS_SM method is introduced in Section IV. The experimental results are shown and discussed in Section V. Finally, the conclusion is drawn in Section VI.

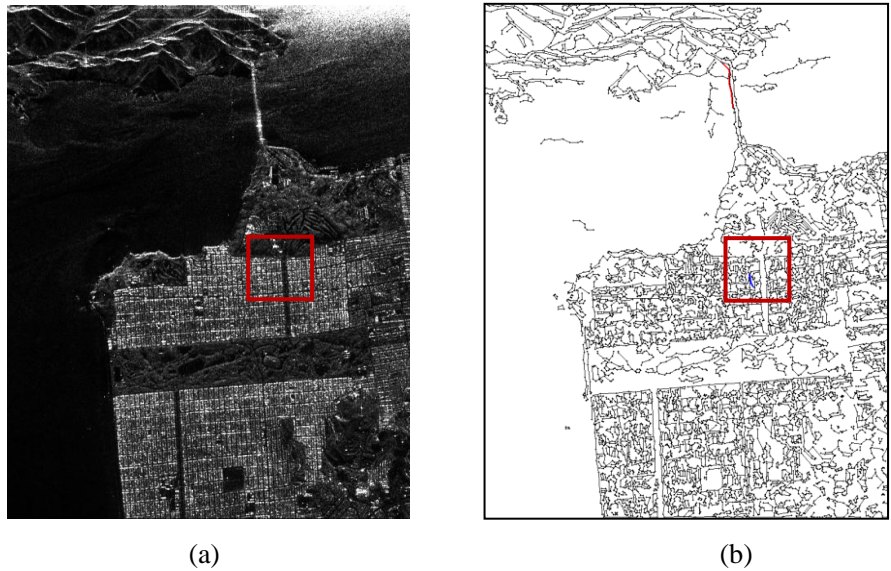

Fig. 1. Example of the aggregated terrain type in low-resolution PolSAR image. (a) SPAN image of San Francisco. (b) Polarimetric sketch map. (c) The building unit and the urban area in (a). (d) Polarimetric sketch map of (c). (e) Sketch segments and their corresponding locations in SPAN image.

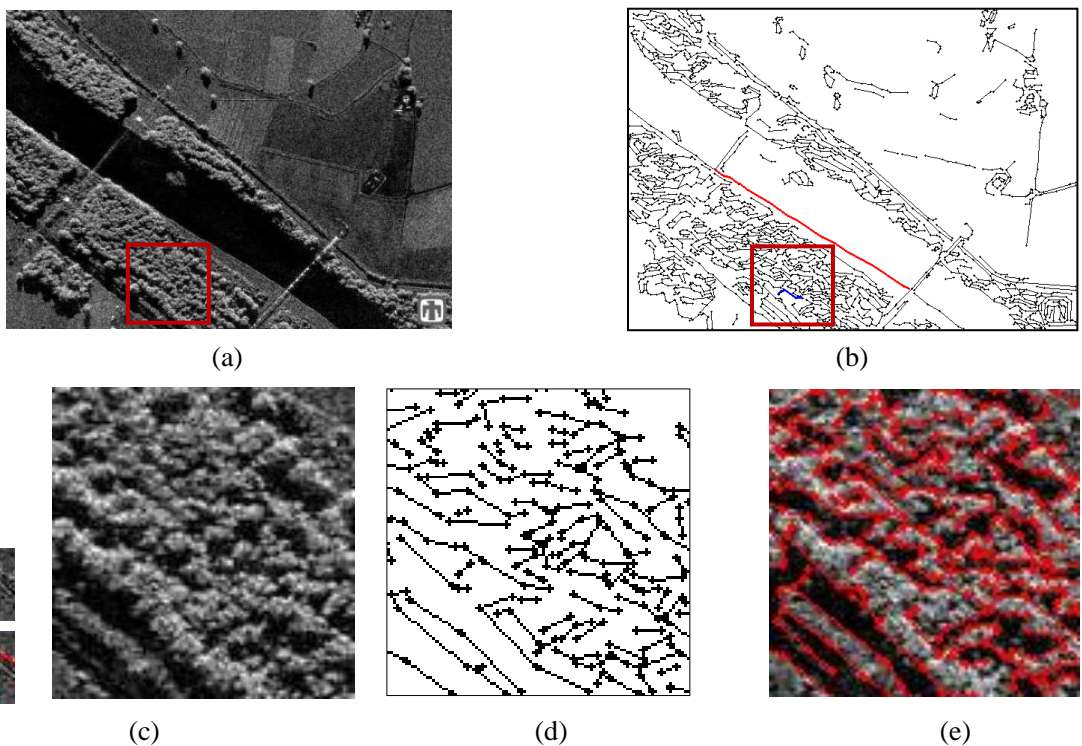

Fig. 2. Example of the aggregated terrain type in high-resolution SAR image. (a) High-resolution SAR image of PIPERIVER. (c) A tree and a forest in SAR image. (d) Sketch map. (e) Sketch segments and their corresponding locations in SPAN image.

## II BACKGROUND AND MOTIVATIONS

In this paper, the relative vision computing theory and the motivations of the proposed method are described in detail below.

## A. Vision Computing Theory

Integrating the psychophysics, neurophysiology and physiological experiments and the computer science, Marr proposed the vision computing theory [32] which has produced a profound influence on artificial intelligence and neuroscience. In the theory, vision perception is considered as an information-processing task which is implemented by constructing a set of representation. From the framework shown in Fig. 3, we can see the representation of an image contains three levels: primal sketch, 2.5 dimensional sketch and 3D model. Primal sketch is a symbolic representation in the low-level vision, and it is useful for high-level process.

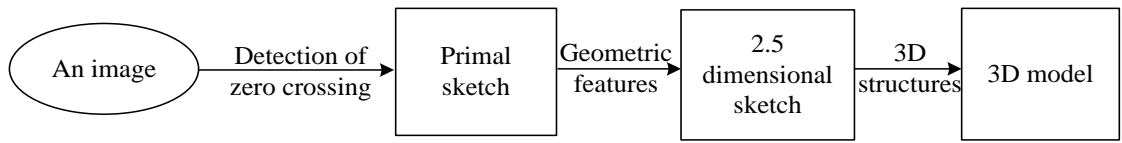

Fig. 3. Three level representations in vision system.

Following Marr's primal sketch model, Zhu et al. [27] gave a mathematical formulation and model schemes to partition an image into sketchable and non-sketchable parts. In the primal sketch model, a sketch graph is extracted as the sparse representation of an image. The procedure of the sketch graph extraction is shown as follows:

1) Detect edges and lines by Gaussian filters with multiple scales and orientations, and obtain the edge map by the non-maxima suppression algorithm.

2) Derive the primal sketch graph by the greedy sketch pursuit algorithm and selection of sketch lines.

The sketch graph, which consists of sketch segments, describes the basic structure of an image.

As a sparse image representation, the primitive in the sketch graph is a set of meaningful sketch segments rather than pixels. The primal sketch model can reduce the additional noise effectively by using Gaussian filters.

## B. Motivations

The PHSM consists of two levels: the polarimetric sketch map and the region map. In this section, the motivations of constructing the two levels are given respectively.

### a) Motivation of Polarimetric Sketch Map

The SAR and PolSAR images have unique characteristics include: shadow, layover, shrinking and shift, since the imaging mechanism is totally different from natural images. Moreover, the aggregated terrain types will formulate sharp bright-dark intensity variations due to the imaging mechanism. Some region-based methods without semantic information have difficulty in representing the aggregated objects, which results in some over-segmented regions in aggregated terrain types.

Marr's vision theory [32] gave a framework of how to process an image like human's vision system and proposed a primal sketch model. The sketch graph in natural image is caused by intensity variations which usually describe the edges. Considering the imaging mechanism and speckle noises of PolSAR images, we propose a polarimetric sketch map as a sparse image representation. By extensive experiments, we found the sketch segments in the polarimetric sketch map have two semantic meanings: 1) edges or line objects; 2) aggregated ground objects.

*Edges or line objects:* Edges or line objects preservation is the main criterion during image classification. When some sketch segments are laid end to end, they will formulate a sketch line. It is found that isolated sketch segments or long sketch lines with linear rank can describe edge or line objects. Take the polarimetric sketch maps in Fig. 1(b) and Fig. 2(b) as examples. The red sketch lines which are long straight lines represent the bridge and the edge of the river respectively.

*Aggregated ground objects:* Sketch segments can also describe aggregated ground objects. It is a sparse symbol representation of aggregated ground objects. Thus, the aggregated region can be formulated by exploiting the spatial relationship of these sketch segments.

Take a low-resolution PolSAR image as an example. Fig. 1(c) is the building and the urban area of the SPAN image in San Francisco. Due to different scattering echoes, the building and its surrounding ground objects formulate sharp bright-dark variation in intensity especially for the building and the thin road. Thus, a sketch segment is formulated to represent the building and its close ground.

In addition, sketch segments can also describe aggregated objects in high-resolution SAR images as shown in Fig. 2. Fig.2 (c) indicates a tree and a forest in a high-resolution SAR image. It can be seen the tree in the forest formulate the bright-dark variation due to the tree and its shadow. Fig. 2(d) is the sketch map of (c) and (e) is the corresponding locations of sketch segments. It can be seen multiple of sketch segments are extracted in the forest due to the trees aggregated together. Thus, sketch segments can represent the aggregated object due to the intensity variation.

## b) Motivation of Region Map

For PolSAR scene classification, one of the difficulties is hybrid terrain types with heterogeneous structures in a PolSAR image. These terrain types are with various scales and orientations and difficult to classify them well by a unified model. Take Fig. 1 as an example. There are the urban area, the sea, the mountain and the forest in this PolSAR image. These terrain types have great differences in structure and are hardly to be segmented into semantic homogenous regions.

By intensive experiments, we found that the spatial distribution of sketch segments can indicate the difference in structure for various terrain types. From Fig. 1(b) we can see that there are almost no sketch segments in the sea while sketch segments in the urban area are aggregated. In addition, the sketch segments in the bridge are non-aggregated and linear rank. So, can we extract semantic homogenous regions as a sparser image representation from the polarimetric sketch map by exploiting spatial relationship of sketch segments?

It is obvious we can extract three parts from a complex PolSAR scene according to the distribution of sketch segments, and each part has a relatively coherent structure. One is the non-sketchable regions which represent homogenous terrain types such as the sea and the farmland. The second is the aggregated regions with aggregated sketch segments, such as the urban area, the forest. And the last one is the structural regions with other sketch segments, such as the edge of two terrain types and the line objects.

Hence, how to obtain semantic homogenous regions in pixel space can be transformed into how to characterize the spatial relationship of the sketch segments and group them on the polarimetric sketch map. According to the aggregated characteristics and spatial rank of sketch segments, a region map can be constructed to partition a complex PolSAR scene into aggregated, structural and homogenous regions. Then, specific classification models can be designed for different regions.

# III CONSTRUCTION OF POLARIMETRIC HIERARCHICAL SEMANTIC MODEL

Our main idea is to partition a complex PolSAR scene into several structural relatively coherent regions and apply suitable segmentation and classification methods to each region. The framework of the proposed approach is shown in Fig. 4. To generate a region partition of the PolSAR image, we construct a PHSM which consists of primal-level and middle-level semantic models, and further a polarimetric hierarchical semantic space is formulated. Based on semantic models, the PolSAR image is partitioned into aggregated, structural and homogenous subspaces. Then, different segmentation or classification methods are applied to the three subspaces. In addition, the segmentation or classification map can further modify the semantic models in the hierarchical semantic space. In this section, we introduce the proposed PHSM by describing the two levels in detail.

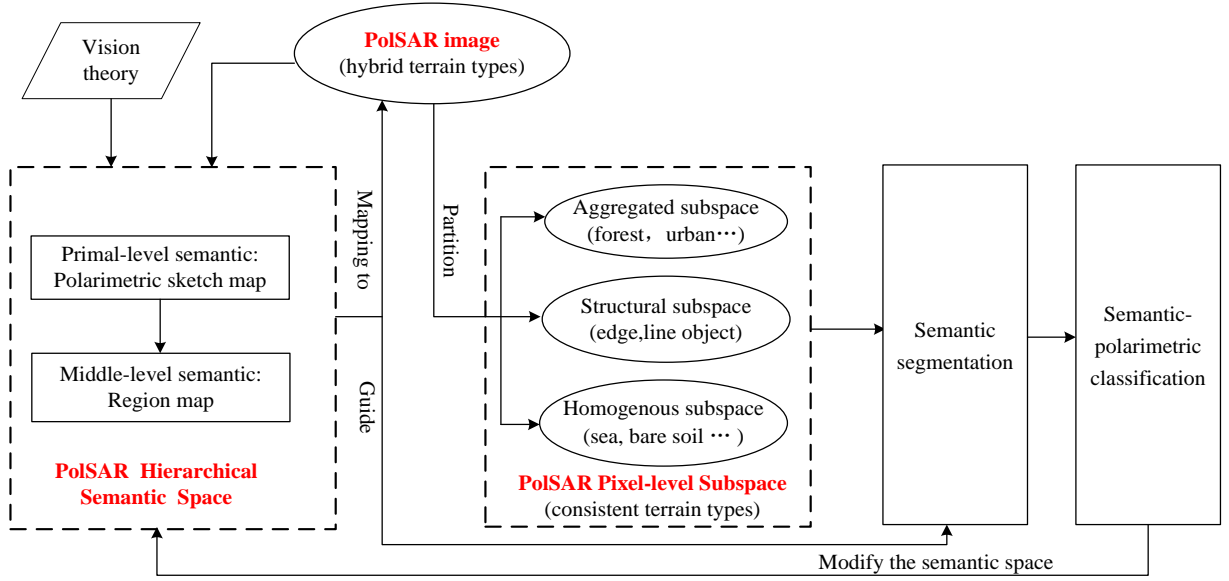

Fig. 4. Framework of the hierarchical semantic model-based classification approach.

## A. Primal-level Semantic: Polarimetric Sketch Map

Considering polarimatric information, a polarimetric primal sketch model is formulated for PolSAR images. There are mainly two steps which are totally different from the primal sketch model in natural images. One is polarimetric hybrid edge-line detection and the other is selection of sketch lines. The main procedure is illustrated in Algorithm 1.

---
Algorithm 1: Procedure of Extracting Polarimetric Sketch Map
---

Step 1: *polarimetric hybrid edge-line detection*: a polarimetric weighted CFAR edge-line detector is designed for the PolSAR data and a weighted gradient method is proposed to enhance the edges in heterogeneous regions. Then, a fusion operator is used to obtain the final edge map.

Step 2: *Selection of sketch lines*: greed primal pursuit is applied to formulate the sketch lines and the sketch lines are selected by a hypothesis-testing method.

---

***Polarimetric hybrid Edge-line Detection:*** A polarimetric hybrid edge-line detection method is proposed to consider both polarimetric characteristics and image information. Firstly, considering the speckle characteristics, the constant false-alarm rate (CFAR) weighted edge-line detection method is proposed for edge-line detection. The CFAR detection method [34] can make full use of polarimetric information and suppress the speckle noise by considering the statistical distribution. However, the homogenous assumption in a filter does not hold on any more in heterogeneous regions such as the urban area. The edge in the heterogeneous region cannot be detected well even though there are sharp bright-dark variations in intensity. In this paper, an anisotropic Gaussian kernel [33] is used to weight the filter for reducing the heterogeneity. In addition, since the ground objects in PolSAR images are with various scales and orientations, filters with multiple scales and orientations are constructed for edge and line detections of PolSAR images. The weighted CFAR edge and line energies of a pixel are given by:

$$E_{edge} = -2\rho \log Q_{12} \qquad (1)$$

$$E_{ridge} = \min\{-2\rho \log Q_{12}, -2\rho \log Q_{13}\} \qquad (2)$$

$$\text{and} \qquad \rho = 1 - \frac{2p^2-1}{6p}(\frac{1}{n}+\frac{1}{m}-\frac{1}{n+n}) \qquad (3)$$

$$Q_{ij} = \frac{(n+m)^{p(n+m)}}{n^{pn}m^{pm}} \cdot \frac{|\bar{Z}_i|^n |\bar{Z}_j|^m}{|\bar{Z}_i+\bar{Z}_j|^{n+m}} \qquad (4)$$

where $Q_{ij}$ is Wishart likelihood ratio of regions $i$ and $j$ [34], $n$ and $m$ are the number of looks of regions $i$ and $j$ in a filter respectively, and $n=m$ in this paper. $p$ is the number of channels, and $\bar{Z}_i$ and $\bar{Z}_j$ is the weighted average of the coherence matrixes in regions $i$ and $j$ respectively. From Eq. (1), we can see that the edge energy is increasing with the decreasing Wishart likelihood ratio. Eq. (2) indicates a line object can be detected when two edges appear with high energy in both sides of the central region. The maximum energy of a PolSAR image is obtained by comparing the $E$ in each scale and orientation.

Since the difference of the PolSAR data can provide complementary information for heterogeneous regions, gradient weighted detector for PolSAR data is proposed to enhance the edge-line energy in the urban area. Specifically, the coherency matrix of PolSAR data is vectored and the Euclidean distance is used to measure the vector difference. The anisotropic Gaussian kernel is used to weight the filter. Furthermore, Since the PolSAR data vary dramatically and are mostly close to zero, a logarithmic transformation is applied to reduce the variation. The gradient weighted edge and line energies of a pixel are given by:

$$G_{edge} = \log \| \sum_{i=1}^{n} w_i V_i - \sum_{j=1}^{m} w_j V_j \|_2 \qquad (5)$$

$$G_{ridge} = \min(G_{edge}^{1/2}, G_{edge}^{1/3}) \qquad (6)$$

where $V_i$ is the vector of the coherency matrix, $w_i$ is the anisotropic Gaussian kernel.

The fusion scheme in [33] is applied to fusion the polarimetric CFAR and gradient weighted edge-line since it can produce high edge-line energies in both weak edges and sharp intensity variations of heterogeneous regions.

*Selection of Sketch Lines*: Similar with the primal sketch model [27], the greedy primal pursuit method is applied to the edge map to formulate the sketch lines. A sketch line consists of several head-tail adjacent sketch segments. Then, the significance of sketch lines are calculated by a hypothesis-testing method [33] which tests the homogeneity between both sides of the sketch lines with Wishart distribution. Thus, the sketch lines caused by speckle noises can be removed under a coding length gain (CLG) threshold which can be selected adaptively by the histogram of the significances of lines. Generally, the fist peak of the histogram is chosen as the CLG threshold.

The PolSAR image of San Francisco area is used to illustrate the procedure of the polarimetric sketch model. The pseudo color image of San Francisco in Pauli base is shown in Fig. 5(a). Fig. 5(b) is the polarimetric energy map obtained by weighted CFAR detector. From Fig. 5(b) we can see that the weighted CFAR detector can obtain high energy in edges of two terrain types while the edge energy is low in the urban area. Fig. 5(c) is the energy map by proposed weighted gradient detector.

The fused energy map is indicated in Fig. 5(d). It can be seen that it not only can obtain high energy in weak edges but also enhance the energy of bright-dark intensity variation in urban area. Fig. 5(e) is the energy map by the weighted gradient detector without the logarithmic transformation. As shown in Fig. 5(e), most of the gradient values are too low and only few of pixels have high values. Fig. 5(f) is the energy map by fusing (b) and (e). It indicates (e) is hardly to provide effective edge energy for the urban area due to too low values. So, the logarithmic transformation is necessary. Fig. 5(g) is the final polarimertic sketch map and Fig. 5(h) indicates corresponding positions of sketch segments on the SPAN image. From Fig. 5 (g) and (h) we can see that both the structures of aggregated areas and edges can be represented by sketch segments with length and orientation. The sketch map is a sparse structure representation of the PolSAR image which is consistent with brain's sparse cognition procedure.

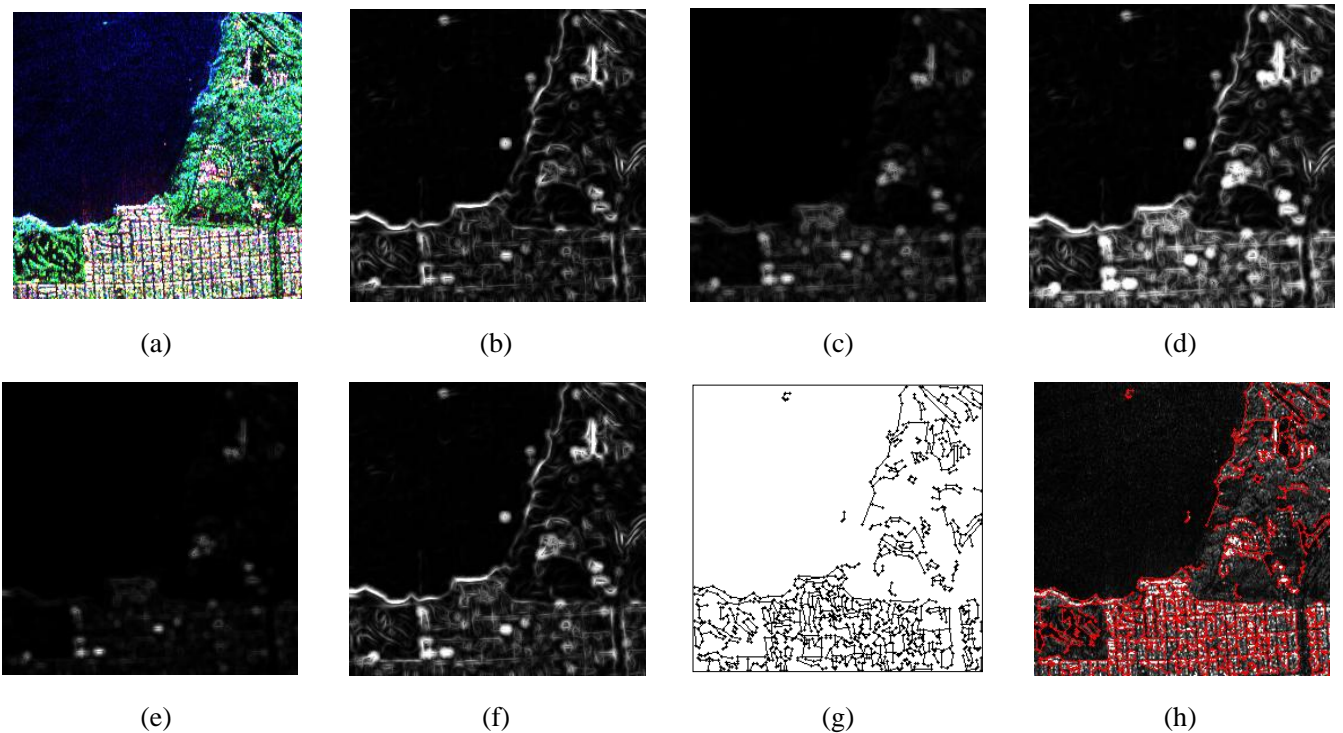

Fig. 5. Example of extracting polarimetric sketch map. (a) PolSAR subimage of San Francisco. (b) Polarimetric energy map by weighted CFAR detector. (c) Gradient energy map. (d) Energy map by fusing (b) and (c). (e) Gradient energy map without logarithmic transformation. (f) Energy map by fusing (b) and (e). (g) Polarimetric sketch map. (h) Corresponding positions of sketch segments on the SPAN image.

Furthermore, two other PolSAR images are used to illustrate the polarimetric sketch maps as shown in Fig. 6. Fig. 6(a) is two PolSAR images in Ottawa and Xi'an respectively. Fig. 6(b) is the energy map by proposed hybrid edge-line detection method. Fig. 6(c) illustrates the polarimetric sketch maps of two PolSAR images. Fig. 6(d) shows the corresponding positions of sketch segments on the SPAN images. It can be seen that the sketch segments can be obtained in both edges and heterogeneous regions with sharp intensity variations.

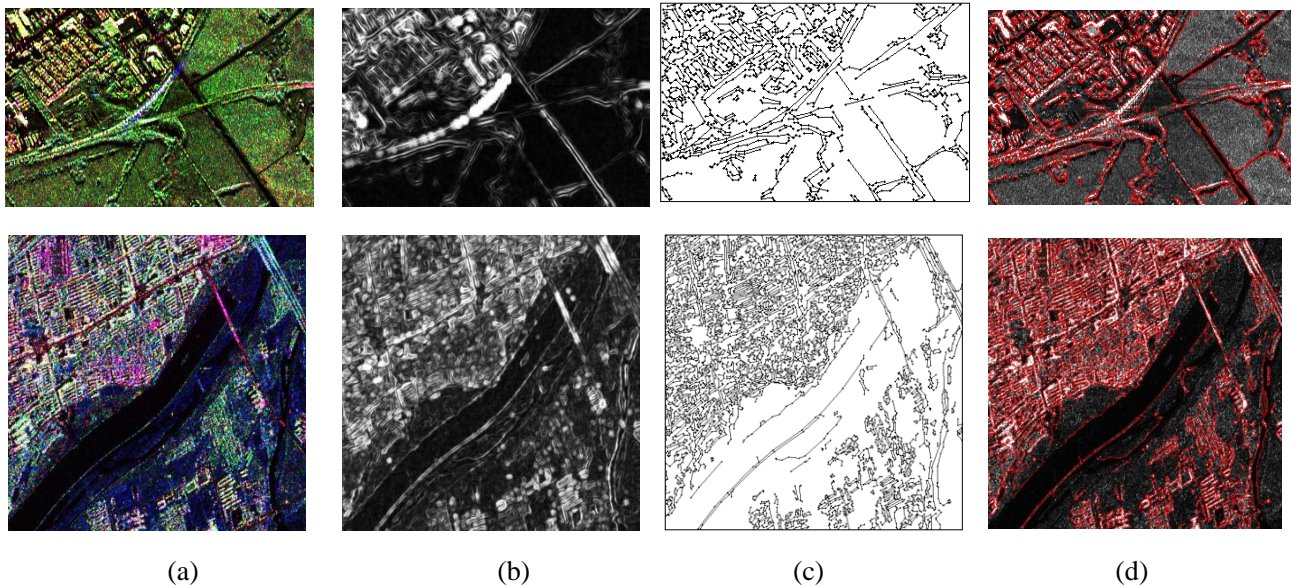

(a)          (b)          (c)          (d)

Fig. 6. Examples of polarimetric sketch map. (a) Two PolSAR images in Ottawa and Xi'an respectively. (b) Energy maps by the proposed hybrid edge-line detector. (c) Polarimetric sketch maps. (d) Corresponding positions of sketch segments on the SPAN images.

## B. Middle-level Semantic: Region Map

Based on the semantic meanings of sketch segments, the region partition of a PolSAR image can be transformed into segment grouping problem on the polarimetric sketch map. The region map algorithm is indicated in Algorithm 2.

| Algorithm 2: Procedure of Region Map |
| --- |
| Step 1: Segment label algorithm via aggregated characteristics and spatial rank. |
| Step 2: Aggregated region extraction method. |
| Step 3: Structural region extraction with geometric structural block. |

### a) Segment Label via Spatial Relationship of Sketch Segments

In this section, we define three characteristics of sketch segments by exploiting their spatial topological structure. Furthermore, based on the three characteristics, sketch segments are labeled as aggregated segments (AS) and isolated segments (IS).

***Continuity of Sketch lines:*** Edges or line objects always consist of one or several head-tail adjacent sketch lines. When head-tail adjacent sketch lines are connected, a longer line is formulated. The long straight lines are considered as the isolated segments. Some definitions are firstly given as follows:

• *sketch segment set*: $S = \{s_i \mid s_i = \{l_i, \theta_i, c_i\}, i = 1 \sim N\}$

where $N$ is the total number of sketch segments, and $s_i$ is the $i$th sketch segment. and $l_i$, $\theta_i$ and $c_i$ represent the length, orientation and central point of $s_i$ respectively.

• *sketch line*: $line_i = \{s_1, s_2, ..., s_k \mid d_1(s_i, s_{i+1}) \leq 2, s_i \in S\}$

where $d_1(s_i, s_{i+1})$ is the distance between the tail of segment $s_i$ and the head of segment $s_{i+1}$ in the line. Due to the speckle effectiveness, the sketch line may be broken with a small distance. In order to connect the broken sketch lines, we set $d_1(s_i, s_{i+1}) \leq 2$.

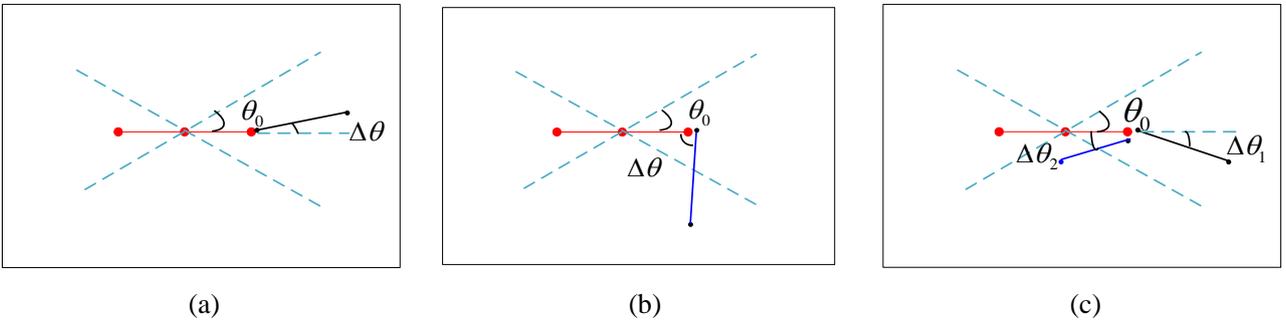

(a) (b) (c)

Fig. 7. Three cases for the continuity of sketch lines. (a) A straight line. (b) An inflect line. (c) Multiple lines. The red segment is segment $s_i$ and the black and blue segments are the sequent segment $s_j$. The black segment can formulate a straight line with $s_i$ and the blue segment will formulate an inflect line.

In a sketch line, there are three cases for the rank of two head-tail adjacent sketch segments: 1) a straight line, 2) an inflict line; 3) multiple lines. We define the orientation variation between two adjacent segments as $\Delta\theta_i = |\theta_{i+1} - \theta_i|$. As shown in Fig. 7, the first case in Fig. 7(a) is a straight line satisfying $\Delta\theta_i \leq \theta_0$ where $\theta_0$ is the threshold of $\Delta\theta_i$. The second case in Fig. 7(b) is an inflect line with $\Delta\theta_i > \theta_0$. The last case is multiple adjacent segments. As illustrated in Fig. 7(c), although both two segments satisfy $\Delta\theta_i \leq \theta_0$, it is obviously the blue segment with $\Delta\theta_2$ has sharp orientation variation and should not be the isolated segments.

Therefore, in order to label the isolated sketch segments, a long straight line is defined as:

$$str\_line_j = \{s_1, s_2, ..., s_k\}$$
$$s.t. \begin{cases} d_1(s_i, s_{i+1}) \leq 2, \\ \sum_{i=1}^{k} l_i > l_0, \\ \max\{\Delta\theta_i\}_{i=1:k} < \theta_0, \\ d_2(s_i, s_{i+1}) > \max\{l_i, l_{i+1}\}, i = 1, ..., k-1 \end{cases} \quad (7)$$

where $l_0$ is the length threshold, $l_i$ is the length of the $i$th sketch segment, and $d_2(s_i, s_{i+1})$ is the distance of the head of $s_i$ and the tail of $s_{i+1}$. $d_2(s_i, s_{i+1}) > \max\{l_i, l_j\}$ guarantees the sketch line is forward and not goes back. In order to consider the edges with slow orientation variation such as the river, we take $\theta_0 = 30°$. In addition, in order to avoid uncorrected label, we only take 5% of the total straight lines as isolated segments after sorting the length in descend. Hence, $l_0$ is selected adaptively by the proportion.

The selection of long straight lines can be shown in Fig. 9. Fig. 9(a) is the SPAN image of Ottawa area and Fig. 9(b) is the polarimetric sketch map. In Fig. 9 (c), the red segments are the labeled isolated segments according to the long straight lines. It can be seen that some long edge or line objects has been correctly labeled as isolated segments. Also, there are still some short edge or line objects which cannot be divided into isolated segments by long straight lines. Other spatial relationships can be involved for better labeling sketch segments.

***Aggregated Characteristics of Sketch Segments:*** By extensive experiments, we found sketch segments in aggregated terrain types are aggregated and sketch segments in edge or line objects are isolated. In this paper, aggregated degree is defined to label sketch segments based on gestalt principles [35-36].

The aggregated degree of sketch segments is proposed to measure the aggregation characteristic of sketch segments. The *K*-nearest neighborhood method is suitable to define the aggregation degree (AD) of a sketch segment since it describes the local aggregated relationship between the sketch segment and its nearest *K* sketch segments. Thus, for segment $s_i$, its aggregation degree *aggregation*(*i*) is defined as:

$$aggregation(i) = \frac{1}{k}\sum_{j=1}^{k} d(s_i, s_j) \quad (8)$$

where $d(s_i, s_j)$ is the distance between segment $s_i$ and its neighbor $s_j$, and the distance of two segments is defined as the Euclidean distance of their midpoints. *k* is the number of the nearest neighbors. The AD can characterize the local structure of a sketch segment and represent the spatial relationship of aggregated ground objects well.

*Spatial Rank of Sketch Segments:* The spatial rank of sketch segments can also characterize different terrain types. According to the spatial rank, the segments can be divided into three types: double-side aggregation segments (DAS), single-side aggregation segments (SAS) and zero-aggregated segments (ZAS)). Zero aggregated segments are considered as isolated segments. DAS is the segment whose close neighbors are located in its both sides. SAS is the segment whose close neighbors are mostly located in only one side. Generally, the DAS and SAS are respectively the inner and boundary segments of aggregated terrain types, thus SAS are labeled as IS. As shown in Fig. 8, the red segment is segment *s*. Fig. 8(a) shows that *s* is a DAS. Fig. 8(b) shows that *s* is an SAS and Fig. 8(c) shows that *s* is a ZAS. It is noted that we did not consider the segments in region 1 when the *k*-nearest neighbors of *s* are computed. Since the segments in region 1 is regarded as approximately parallel to *s* and parallel lines probably are line objects or the boundaries. In order to keep these line objects from merging, the parallel lines should be IS. The region 1 is formed by two straight lines which have the same intersection angle α with segment *s*. Generally, α is taken as a small value, for example, $\alpha = 10°$.

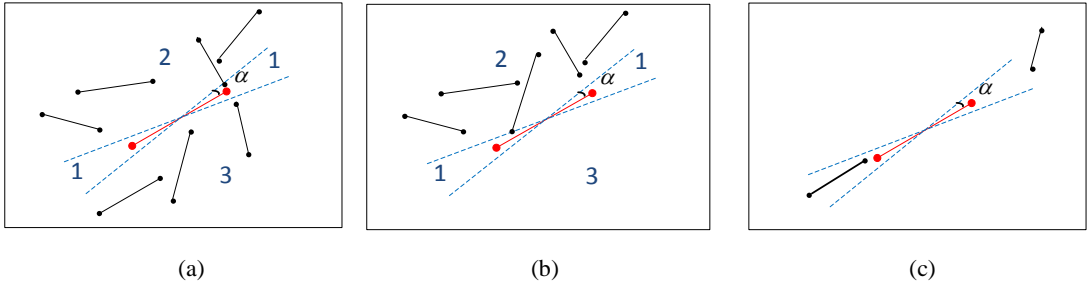

(a) (b) (c)

Fig. 8. Three types of spatial rank for sketch segments, the red one is segment *s* and the black ones are its surrounding segments. According to the distribution between segment *s* and its *K*-nearest segments, segments are divided into three types: (a) double-side aggregation (DAS); (b) single-side aggregation (SAS); (c) zero aggregation (ZAS).

*Segment Label:* According to the aggregated degree, sketch segments can be divided into two types (aggregated segments (AS) and isolated segments (IS)) by a threshold $\delta_1$. We found the aggregation degree histogram of sketch segments (ADH) can guide the selection of $\delta_1$. Take Fig. 9(a) as an example. The ADH of the sketch segments when *k*=9 is shown in Fig. 9(d). The ADH forms a curve with a high peak and a long tail. Generally speaking, the number of AS is much more than IS's thus form a high peak, and the IS is the tail of the ADH. According to the ADH, $\delta_1$ is adaptively determined by:

$$\sum_{\min t}^{\delta_1} p(t) = r \sum_{\min t}^{\max t} p(t) \qquad (9)$$

where $p(t)$ is the frequency of the ADH when AD is *t*. *r* is a ratio of the number of AS to the total segment number and is usually taken as 90-95% for reference, since the sketch segments in aggregated terrain types are much more than that in edges or line objects. Fine adjustments are needed according to specific images.

In addition, the SAS is labeled as IS according to the spatial rank. After segment label, the semantic sketch map is obtained and shown in Fig. 9(e). The blue sketch segments represent AS, and the red sketch segments represent IS. Moreover, the segment label method is indicated in Algorithm 3.

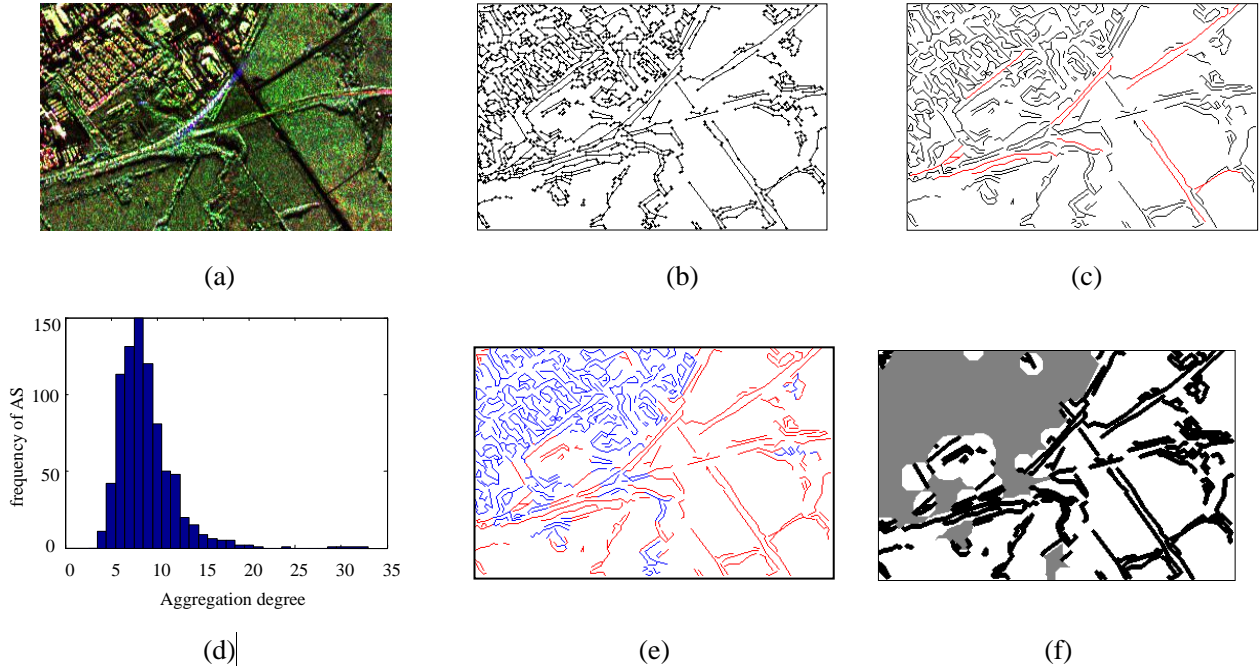

Fig. 9. Example of the region map extraction. (a) SPAN image of Ottawa area. (b) Polarimetric sketch map. (c) Label IS in red with long straight lines. (d) Aggregation degree histogram of sketch segments with $K$=9. (e) Semantic sketch map, the sketch segments in blue represent AS, and the sketch segments in red represent IS. (f) Region map. The gray, black and white regions represent aggregated, structural and homogenous regions respectively.

| Algorithm 3: segment label algorithm |
| --- |
| Step 1: Connect sketch lines and compute the length of sketch lines and variation orientations of two adjacent segments. |
| Step 2: Label long straight sketch lines as IS. |
| Step 3: Calculate aggregated degree of the remaining sketch segments. |
| Step 4: Select the threshold according to the ADH of aggregated degree, and label the remaining segments as AS and IS by the threshold. |
| Step 5: According to the spatial rank, the aggregated segments are divided into DAS and SAS, and SAS are labeled as IS. |

## b) Aggregated Segment Grouping based Region Extraction

According to the segment label, segments are divided into AS and IS. When two sets of AS are far away, they should be grouped into different clusters. Since there are multiple aggregated areas in a PolSAR image, the AS can be grouped into several clusters and each cluster can formulated an aggregated region. Therefore, the extraction of aggregated regions mainly includes two stages. Firstly, the aggregated segment set is grouped into several sets by the segment grouping algorithm. The aggregated segments in a same set should satisfy certain spatial constraints. Then, an aggregated region is obtained with a minimized closed region covering the set of aggregated segments. Some symbols are defined in advance in Appendix 1. The pseudo code of aggregated region extraction algorithm is given in Appendix 2.

**Stage 1: Spatial Constraint Based Segment Grouping Algorithm**

The aggregated segment grouping problem is to partition $U$ into several subsets $\{T_k\}, k = 1, 2, \cdots, m$ which can be described as:

$$T_k = \{s_1, s_2, \cdots, s_{n_k}\}, \quad k = 1, 2, \cdots, m$$

$$s.t. \begin{cases} \bigcup_{i=1}^{m} T_k = U \\ T_i \cap T_j = \varnothing \\ \forall s_i \in T_k, \exists s_j \in T_k \quad s.t. \quad d(s_i, s_j) \leq \delta_2, i \neq j \end{cases} \quad (10)$$

where $n_k$ is the segment number in $T_k$, and $d(s_i, s_j)$ is the distance between segment $s_i$ and segment $s_j$.

A segment grouping algorithm is proposed to calculate the aggregated segment subsets. This method is similar to the region growing algorithm but with the sketch segment as a unit and sketch segments sets as a result. The segment grouping criterion is the spatial constraint threshold $\delta_2$.

The spatial constraint threshold $\delta_2$ is adaptively selected according to the ADH. Since the peak of ADH generally reflects the average AD in aggregated regions, threshold $\delta_2$ should be slightly larger than the peak value to ensure that the sketch segments in an aggregated area are gathered together. Generally, we set $\delta_2$ as the average AD of sketch segments for reference. The threshold $\delta_2$ is defined as:

$$\delta_2 = \frac{1}{N} \sum_{i=1}^{N} t \quad (11)$$

where $t$ is the AD of the $i$th sketch segment and $N$ is the total number of segments. The threshold $\delta_2$ can also have a fine adjustment for a specific image.

**Stage 2: Aggregated Region Extraction Based on Groups**

Since the aggregated segment stands for the intensity variation between objects and the earth or their shadows in aggregated land cover, the set of the aggregated segments formulates an aggregated land cover. With the aggregated segment subsets, aggregated regions can be extracted by finding a close region to fulfill the gaps of the aggregated segment subset. The morphological closing operation is an effective tool to obtain closed regions on an image using a structural element. Therefore, in order to obtain a homogenous region in an aggregated ground object, the morphological closing operation is used in each aggregated segment subset with a circle element. And the circle element with the radius of $\delta_2$ is selected to fill the biggest gap within the aggregated sketch segment subsets.

### c) Structural Region Extraction

In order to preserve the edges or line objects and further find the real boundary, the structural regions, should be extracted from IS. A geometric structural block [33] is applied to extract the regions of edges or line objects. And its width is usually taken as three or five since the line object only takes several pixels in width. The remaining regions of an image are homogenous regions.

As indicated in Fig. 9(f), a region map is achieved which partitions an image into three parts: aggregated, structural and homogenous regions colored by gray, black and white respectively. As the middle level semantic, the image primitive is the meaningful region exploited from the primal level semantic space. The region map not only merges the aggregated terrain into a region but also keeps the edge and line object regions.

The proposed PHSM is illustrated by three PolSAR images in Fig. 10. The first column is Pauli image of PolSAR data from different sensors and bands, and the second column is corresponding SPAN images. The common characteristic of these PolSAR images is that scattering waves of the urban area formulate bright-dark variation in intensity by the building and the earth. The third column is polarimetric sketch maps. The sketch segments in the sketch map represent the structures of the image. According to the spatial relationship of sketch segments, a region map, as shown in the fourth column, is extracted from the sketch map. The region map partitions an image into aggregated, structural and homogenous regions in gray, black and white respectively.

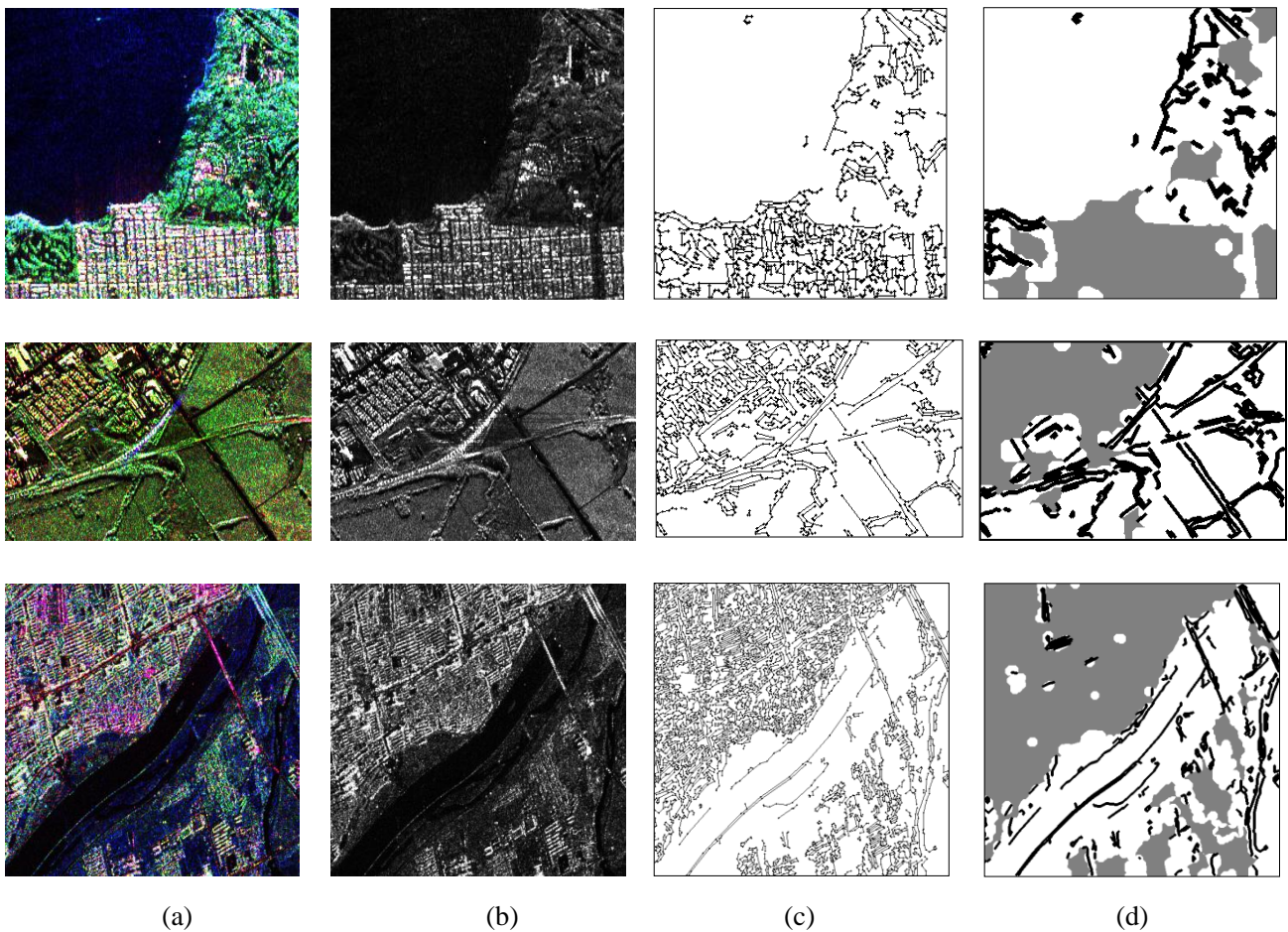

(a)      (b)      (c)      (d)

Fig. 10. Examples of the proposed polarimetric hierarchical semantic model. (a) Three PolSAR images with Pauli base as RGB color channels. (b) SPAN images. (c) The primal-level semantic: sketch maps, and (d) the middle-level semantic: region maps.

## IV POLSAR TERRAIN CLASSIFICATION BASED ON THE PHSM AND SCATTERING MECHANISM

Based on the PHSM, a PHS_SM method is proposed for POLSAR data by exploiting different characteristics of three regions. The procedure of the proposed PHS_SM method is illustrated in Algorithm 4. The semantic segmentation and the semantic-polarimetric classification will be introduced in detail next.

| Algorithm 4: the proposed PHS_SM method |
| --- |
| Step 1: the PHSM is constructed with Algorithm1 and 2. |
| Step 2: Semantic segmentation based on the PHSM. |
|     1) Initial partition by mapping the region map to the over-segmented map. |
|     2) Region merging in aggregated subspace. |
|     3) Edge location in structural subspace. |
|     4) Hierarchical segmentation in homogenous subspace. |
| Step 3: Semantic-polarimetirc classification. |

## A. Semantic Segmentation based on the PHSM

Mapped by the region map, a PolSAR image is partitioned into three subspaces. Considering different characteristics in three subspaces, we design three different merging schemes based on an initial over-segmentation map.

*Initial partition:* Superpixels, which are the over-segmented regions, can not only reduce the computational time but also produce more homogenous regions. A mean-shift algorithm [37] is used for initial segmentation for its less over-segmented regions and better edge preservation.

As a crude image partition, the region map can guide the merging of the superpixels. The region map is mapped to the superpixels which are divided coarsely into three parts. The first part is superpixels corresponding to aggregated regions. The second part is superpixels corresponding to structural regions, and the last part is other superpixels. Three distinct merging schemes applied to three parts are described in detail next.

*Merging Scheme in Aggregated Subspace:* The aggregated region is semantic homogenous while the boundaries are not accurate. On the contrary, considering the image information, superpixels have real boundaries while too many over-segmented regions in the inner of the aggregated terrain types. A majority voting scheme [38] is applied to merge the aggregated regions as shown in Fig. 11. By this, the inner of aggregated regions and can be merged directly and the edge is located by superpixels.

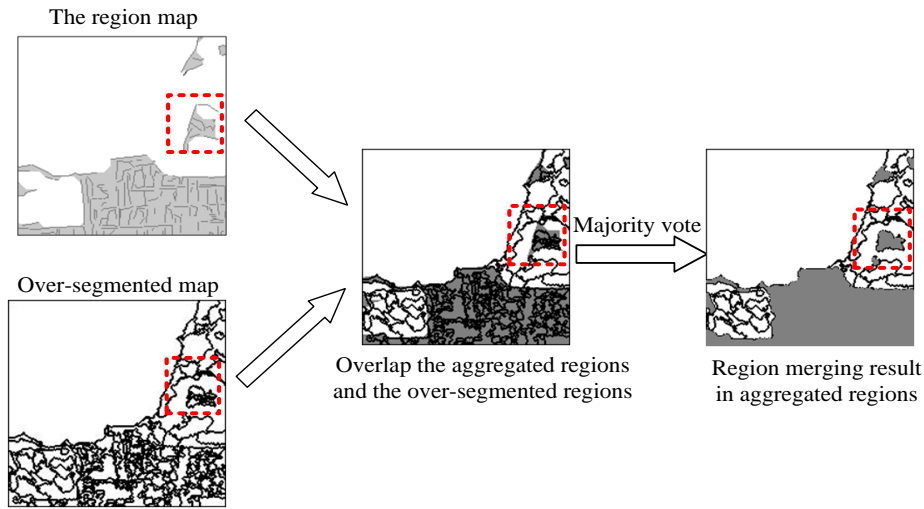

Fig. 11. Example of the merging procedure of aggregated regions for the subimage of San Francisco.

*Edge Location in Structural Subspace:* According to the semantic meanings of sketch segments, the isolated segments represent the line object or the boundary of two different regions. For each structural region, the true boundaries should be located no matter the sketch segment is an edge or a line object. Specifically, to refine the edge-line regions, the true edge is located from the edge map by the polarimetric edge-line detection method, and each structural region is partitioned into two regions along the true boundary. Fig. 12 indicates the edge location in structural regions. The red rectangle in Fig. 12(a) is the extracted structural region and the green line is the sketch segment. The blue curve in Fig. 12(b) is the located edge which partitions the aggregated region into two parts.

*Hierarchical Segmentation in Homogenous Subspace:* A hierarchical merging method [7] is applied to the homogenous regions. According to the Wishart distribution of the coherence matrix, a maximum likelihood method [7] is applied to measure the similarity of two superpixels. After calculating the similarity, the two adjacent superpixels with the minimum value of SC are merged for each iteration. A simple stopping rule is defined as the region number $N_r$ after merging.

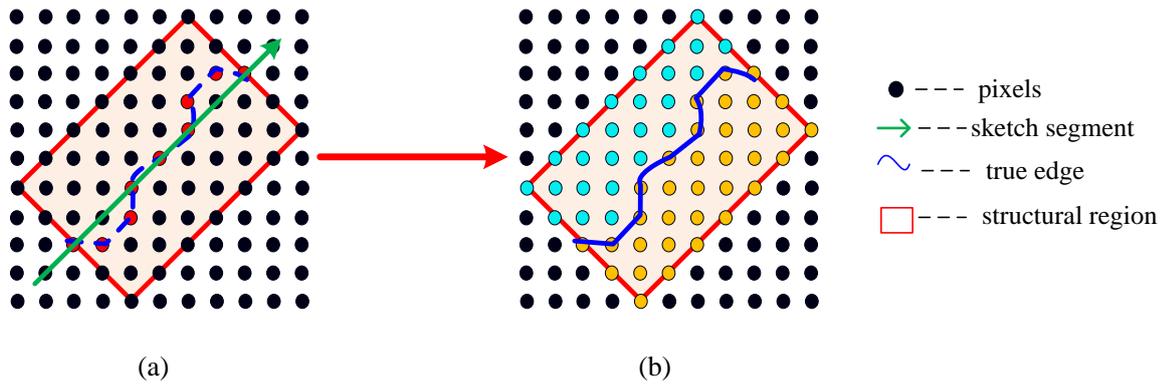

|   (a)   |   (b)   |

Fig. 12. Edge location in structural regions. (a) Structural region extraction. The black points are pixels and the arrow in green represents the sketch segment. The red rectangle is the extracted structural regions and the red points are the true edge. (b) True edge which partitions structural region into two parts. The curve in blue is the true edge obtained by the polarimetric edge-line detection. The points in the structural region are divided into two parts in cyan and orange respectively.

## B. Semantic-polarimetric Classification

Semantic segmentation can provide homogenous regions by exploring hierarchical semantic information. To assign label to each pixel correctly, polarimetric scattering mechanism is utilized for the classification of the POLSAR data. The H/α-Wishart classification method [9] can identify the main terrain types clearly by combining the physical scattering characteristics and statistical properties. However, the region homogeneity is poor without considering spatial information. In order to improve the region homogeneity, a semantic-polarimetric classification method is applied to provide adaptive spatial neighbors for the polarimetric classification.

The semantic-polarimetric classification is based on the majority voting rule [38-39]. Specifically, for each region in the segmentation map, all the pixels are assigned to the most frequent class within this region. The majority voting rule can produce a good classification result with more homogeneous regions by combining the semantic and polarimetric information. The segmentation map can provide an adaptive neighborhood for classification, and the neighborhood of each pixel is defined as the region it belongs to in the segmentation map.

# V EXPERIMENTAL RESULTS AND ANALYSIS

## A. Experimental Data and Settings

We test the proposed PHS_SM method on four sets of the PolSAR data with different bands and sensors: 1) the NASA/JPL AIRSAR L-band data of San Francisco which is a 4-look fully polarimetric SAR data with a dimension of 700×900 pixels; 2) the L-band Oberpfaffenhofen data in Germany, which is provided by the German Aerospace Center's E-SAR; 3) the CONVAIR data of Ottawa Area which is a single-look fully polarimetric complex format data with 222×3429 pixels; and 4) the RadarSAT-2 C-band data of Xi'an Area acquired in China with the resolution of 8m. The ground truth of the second data set is given in this paper and the classification accuracies are illustrated in the experiments.

For all the experiments, the filters with three scales and 18 orientations are selected for edge-line detection because they are sufficient to describe the ground objects. The CLG is selected adaptively by their histogram. $\delta_1$ and $\delta_2$ are calculated according to the Eqs. (9) and (11). All the experiments are conducted on a computer with an Intel core i3 3.20GHz processor and 4.00GB RAM.

Moreover, to verify the effectiveness of our method, we compare the PHS_SM with five related methods including: 1) the common used H/α-Wishart method [7]; 2) Wang et al.'s method [9] denoted by SPECR in this paper, which is an unsupervised classification of fully polarmetric SAR Images based on scattering power entropy and copolarized ratio; 3) the Markov Random Field (MRF) method [40] which involves contextual information into the Wishart classifier; 4) a deep learning method [41] in which high-level features are learned by a deep auto-encoder network followed by a clustering method due to the lack of labeled data; and 5) a method denoted by MSSM, which is the unsupervised classification by only combining the H/α-Wishart classification and the over segmentation from the mean-shift algorithm. The last method is designed to verify the effectiveness of semantic segmentation by using the proposed PHS_SM method without semantic segmentation.

## B. Experimental Results of San Francisco Data

The POLSAR image of San Francisco data is shown in Fig. 13(a) with Pauli base as RGB color channels. The corresponding satellite image from the Google Earth is shown in Fig. 13(b). Complicated ground objects can be seen in this image such as the urban area, the mountainous region, the sea, the vegetation, the bridge and so on.

Fig. 13(c) indicates the region map where the aggregated, structural and homogenous regions are drawn by the gray, black and white colors respectively. From Fig. 13(c) we can see that the urban area is extracted well except for some small refinements. At the same time, different aggregated regions are separated by $\delta_2$. The bridge and some edges and line objects are labeled as structural regions. The mountain ridge is detected as the structural region because the sketch segments in the mountainous region are linear distribution and much sparser than the urban areas'.

Fig. 13(d) indicates the classification result of the PHS_SM method. The classification maps of MSSM, H/α-Wishart, SPECR, Wishart based MRF, and deep learning methods are shown in Figs. 13(e)-(i) respectively for comparison. Parameters in the MSSM algorithm are the same with the PHS_SM method. The SPECR method has the same parameter with [9] in this experiment.

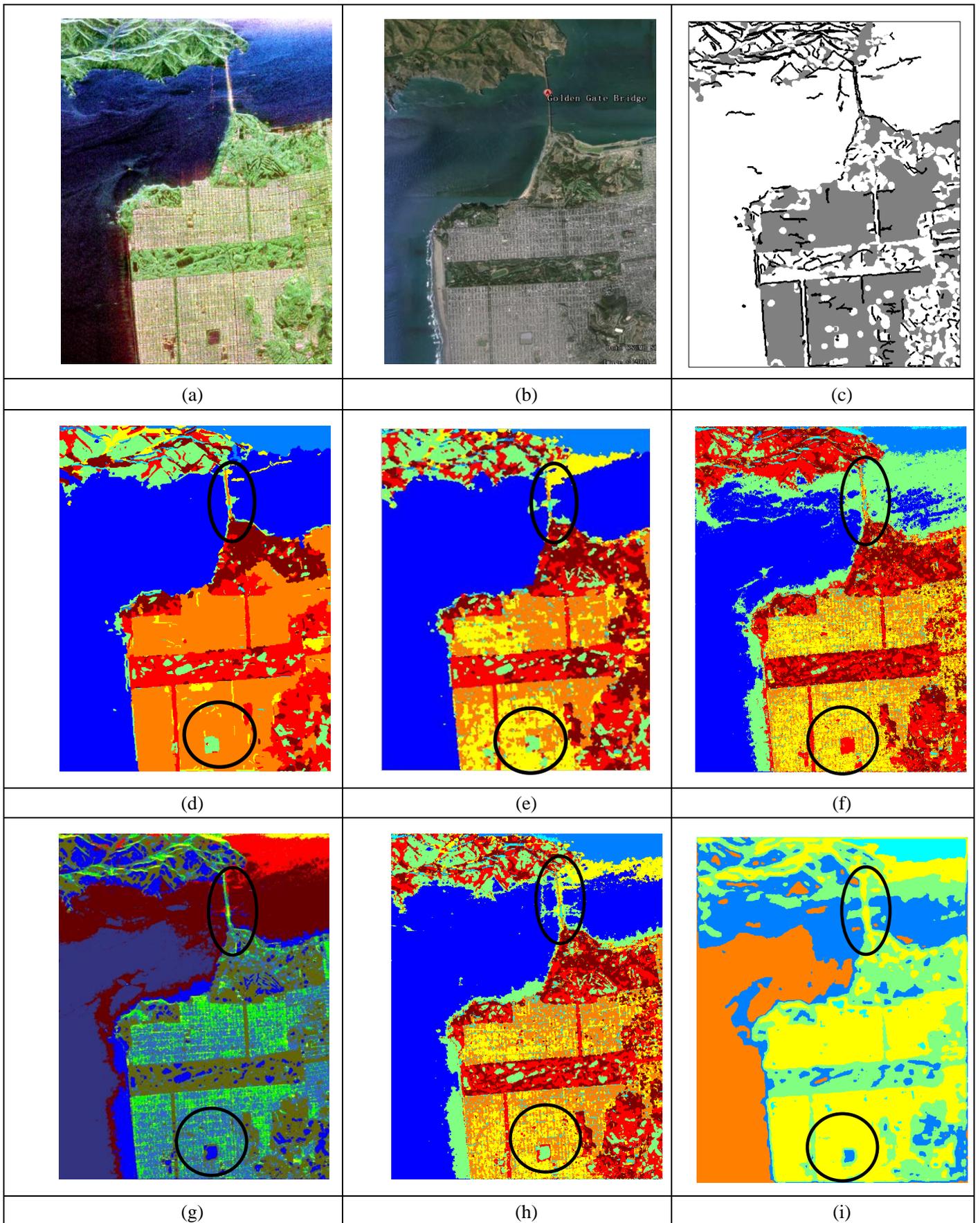

Fig. 13. Classification results of San Francisco using different methods. The circles highlight the details of six results for comparison. (a) Pauli RGB image of San Francisco. (b) Optical image from Google Earth. (c) Region map. (d) Classification by PHS_SM method. (e) Classification by MSSM method which combines the superpixels from mean-shift algorithm and H/α-Wishart classification result. (f) Classification by H/α-Wishart. (g)

Classification by SPECR method. (h) Classification by Wishart based MRF method. (i) Classification by deep learning method.

Although the accuracy rate cannot be calculated due to the lack of ground-truth maps, the optical image from Google Earth is provided for reference in Fig. 13(b). It can be seen from Fig. 13 that our method in (d) can get better performance in region homogeneity and line object preservation compared with classification maps in (e)-(i). The classification result by H/α-Wishart in (f) can obtain accurate pixel label in most pixels by combining the target decomposition and statistic distribution. However, without considering spatial information, this method is sensitive to speckle noise and easily produces a pepper-and-salt classification map. For Fig. 13(e), there are a lot of misclassification especially in the urban area and the bridge which are classified into several classes. The classification result in Fig. 13(g) can get better details but some over segmented regions appear especially in urban area and the sea. Although the classification map by the Wishart base MRF in (h) is slightly better than the H/α-Wishart in (f), it also appears some misclassifications in the urban and the vegetation areas. In addition, although the deep learning method in Fig. 13(i) can show good performance in region homogeneity especially for the urban area, while the edge of the bridge and the mountain cannot be identified clearly. Beside, the labels of the sea and the mountain are confused partly in Fig. 13(i).

The PHS_SM method can identify various classes such as the urban area, forest, water, bridge and grass clearly in Fig. 13(a). Especially in the urban area and the sea, a uniform region is obtained rather than a noised classification map. Semantically homogenous regions are superior for further image understanding. Moreover, the small objects are segmented clearly as shown in the circle for better comparison. It is demonstrated that the PHSM does improve the classification result. However, the beach on the left is missing due to the lack of obvious edges on the image. More effective features will be involved to identity this part in the further work. Consequently, due to the PHSM, the proposed method can show better performance in region homogeneity and line object preservation.

## C. Experimental Results of Oberpfaffenhofen Data

The E-SAR data of Oberpfaffenhofen is also used to validate the effectiveness of the proposed method, which is L-band with 0.92×1.49m pixel size and by a 3.00×2.20 resolution. The PolSAR pseudo color subimage with 212×387 pixels is shown in Fig. 14(a) and the corresponding ground truth is indicated in Fig. 14(b). The optical image from Google Earth is illustrated in Fig. 14(c) for reference. It can be seen that the image is mainly labeled as five classes: *Woodland, Settlement area, Roads, Farmland* and *other* land covers. The ground truth is marked by the optical remote sensing and the label map in [19]. Hence, the ground truth is not exactly accurate which considers the majority terrain type in each class. Furthermore, the *farmland* is ignored during the calculation of the accuracy rate since its area is too small as shown in Fig. 14(b).

The classification maps obtained by the PHS_SM method and other algorithms are shown in Figs. 14(d)-(i). The classification accuracies of these methods are recorded in Table II, and the confusion matrix of the PHS_SM method is listed in Table III.

From Table II and Fig. 14, we can conclude that the PHS_SM method has better performance than other algorithms. Specifically, the average classification accuracy of the PHS_SM method is 71.15%, which is 11.8%, 15.6%, 22.4%, 23.7% and 23.2% higher than other methods respectively. The unclear classifications of the road result in the low accuracy of all the classification methods. As shown in Fig. 14(f), the classification accuracy of *road* is higher than *other* class in the H/α-Wishart method since *other* is misclassified into *road* class. In addition, the *woodland* and the *settlement area* are confused in all the other methods while the proposed method can classify them with both higher accuracies. From Fig. 14 it can be seen that the PHS_SM method obtains more homogenous regions than other approaches. Furthermore, the edge of the woodland can be clearly identified. In the SPECR method, the *settlement area* class is mostly misclassified as the *woodland* thus their classification accuracies are totally unbalanced as shown in Table II. In the deep learning method, *woodland* and *settlement area* classes are confused and the *road* cannot be identified either. From Table III, it is demonstrated that the main reason of the low classification accuracy in the proposed method is the confusion between *road* and *other* classes. It is considered that the Wishart distribution is not suitable for the high-resolution data any more. Better distribution model will be applied in the further work.

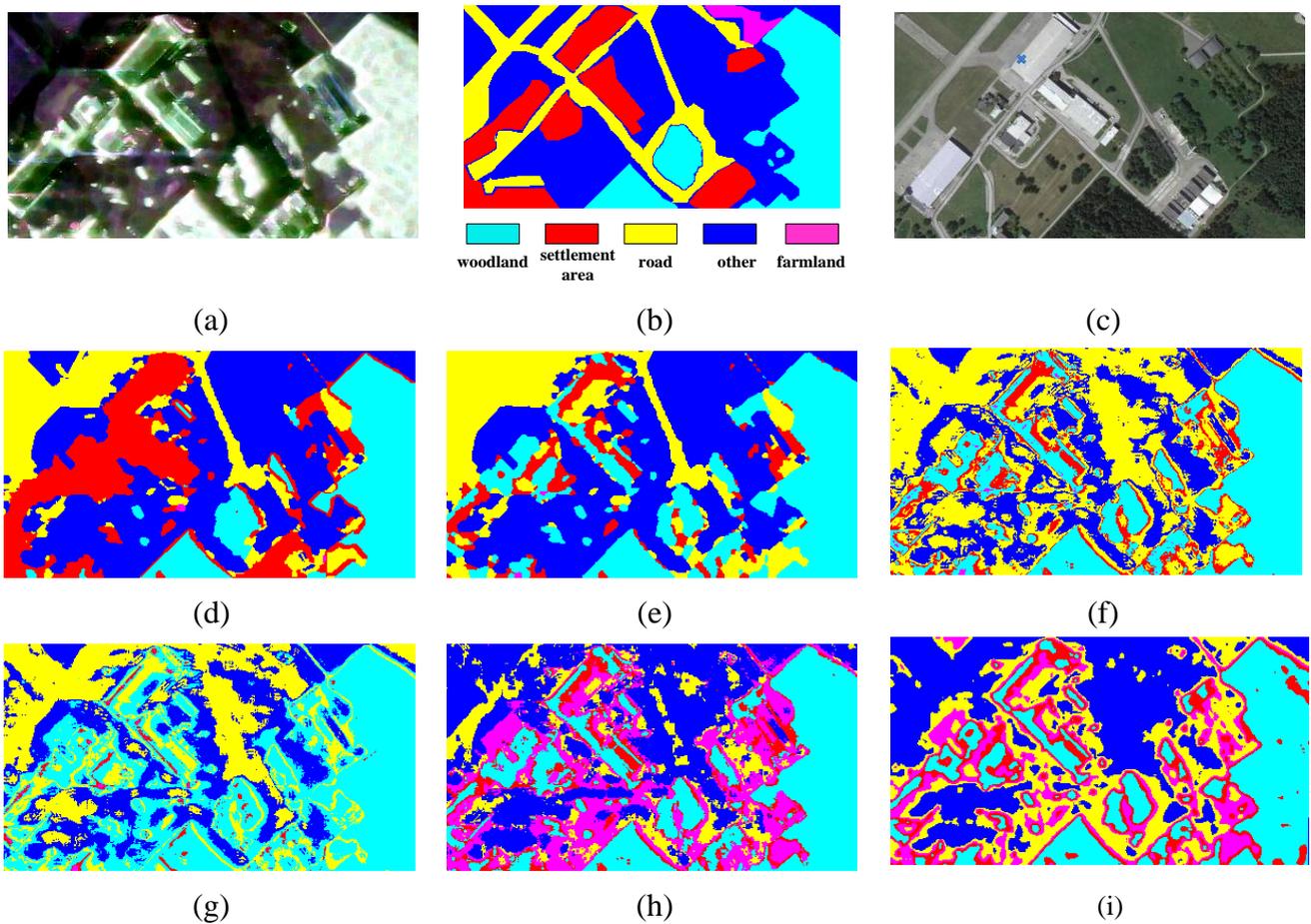

Fig. 14. Classification maps of a sub-region from Oberpfaffenhofen data. (a) Pseudo color image. (b) Ground truth. (c) Optical image from Google Earth. (d) Classification map by PHS_SM. (e) Classification map by MSSM. (f) Classification map by H/α-Wishart. (g) Classification map by SPECR. (h) Classification map by Wishart based MRF. (i) Classification map by deep learning method.

Table II Classification accuracies (%) of six methods

|  | Woodland | Settlement area | Road | Other | Average Accuracy |
|---|---|---|---|---|---|
| PHS_SM | 90.89 | **78.35** | 45.51 | **69.84** | **71.15** |
| MSSM | 92.37 | 27.41 | 49.58 | 67.89 | 59.31 |
| H/α-Wishart | 91.19 | 27.55 | **64.38** | 39.05 | 55.54 |
| SPECR | **92.84** | 6.000 | 56.92 | 39.32 | 48.77 |
| Wishart_MRF | 90.99 | 24.93 | 28.32 | 45.55 | 47.45 |
| Deep learning | 80.24 | 34.13 | 32.07 | 45.25 | 47.92 |

Table III Confusion matrix (%) of the proposed method

|  | Woodland | Settlement area | Road | Other |
|---|---|---|---|---|
| Woodland | 90.89 | 5.190 | 1.280 | 2.640 |
| Settlement area | 1.480 | 78.35 | 5.700 | 14.47 |
| Road | 0.260 | 17.76 | 45.15 | 36.58 |
| Other | 2.330 | 11.43 | 16.31 | 69.84 |

## D. Experimental Results of Ottawa Area Data

A 10-look preprocessing is firstly applied to the original data set of Ottawa Area in azimuth direction. Then the Pauli RGB image with 222×342 pixels is obtained and shown in Fig. 15(a). There are urban area, railway, bare land and roadway in this image. Fig. 15(b) illustrates the SPAN image of Ottawa Area.

The classification results of the PHS_SM method and the other five algorithms are illustrated in Figs. 15(c)-(h) respectively. From Fig. 15, we can see that our method in Fig. 15(c) shows better performance in region homogeneity especially in the urban area than Figs. 15(d)-(h). It indicates the region map is effective for classification of the complex PolSAR scene. The urban area in Fig. 15(c) can be classified into a whole semantic region which is useful for further image understanding. This is explained by the fact that the aggregated region is extracted with semantic information. In Fig. 15(e), the result by the H/α-Wishart method shows a few details but it is greatly affected by speckle noises. The result by the Wishart based MRF method in (g) appears some misclassification in the urban area either. The classification result in Fig. 15(f) can get more homogenous region than Fig. 15(e) while some line objects disappear. The classification map by the deep learning method is good

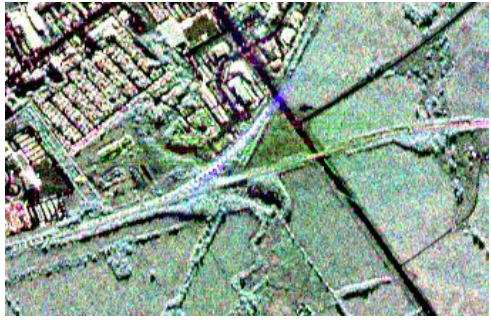
(a)

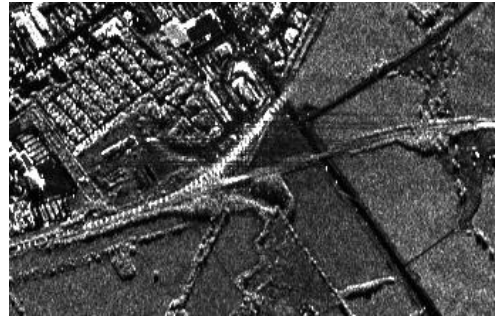
(b)

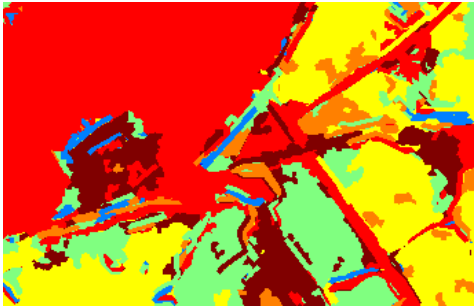
(c)

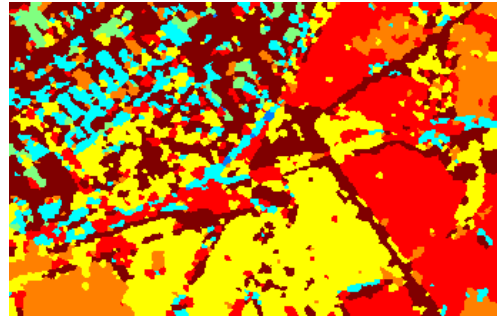
(d)

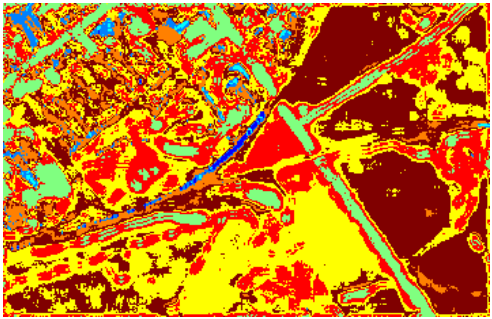
(e)

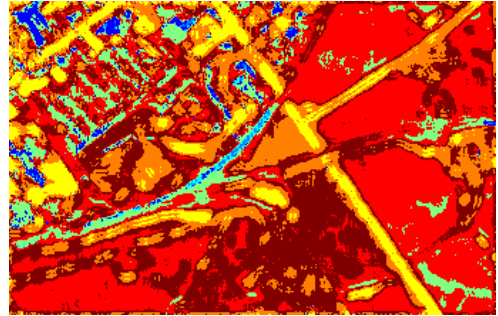
(f)

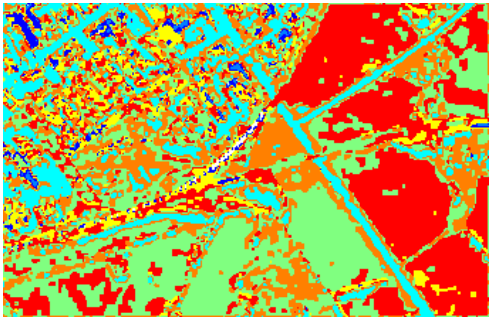
(g)

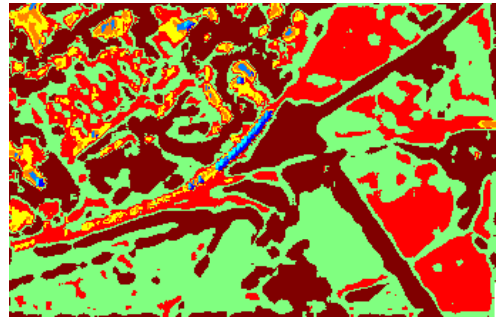
(h)

Fig. 15. Classification results of Ottawa Area. (a) Original image with Pauli base as RGB color channels respectively. (b) SPAN image of Ottawa Area data. (c) Classification by PHS_SM method. (d) Classification by MSSM method which combines the superpixels from mean-shift algorithm and H/α-Wishart classification result. (e) Classification by H/α-Wishart. (f) Classification by SPECR method. (g) Classification map by Wishart based MRF method. (h) Deep learning classification result.

in the region homogeneity but poor in edge-line preservation. The MSSM classification result in Fig. 15(d) without the region map produces lots of over-segmented regions especially in the urban area.

It illustrates that the classification result is greatly affected by the region map, and the PHSM plays an important role in PHS_SM method for the classification.

## E. Experimental Results of Xi'an Area Data

The pseudo color image of Xi'an area is shown in Fig. 16(a). The corresponding optical image from Google Earth is indicated in Fig. 16(b). Since they are obtained in different periods, the optical image is provided for reference. The upper left corner of the image is the urban area and some villages are located on the bottom right corner. A river named Weihe River is in the center of the map and a bridge is across the river. There is a railway parallel to the bridge on the upper right corner of the map.

The classification maps of the PHS_SM method and other five methods are illustrated in Figs. 16(c)-(h). There are mainly four terrain types in this image: the urban area, the villages, the river and the grass. The classification result in (c) show better performance than other methods especially in urban area and the grass. Other methods give rise to some confused classes in these areas even though more details are obtained. In addition, the urban area is classified into homogenous regions in (c) while misclassification appears in most of the methods. Homogenous regions can be achieved in Fig. 16(h) while the edge of the river cannot be identified especially on the right of the image. From Fig. 16(c) it can be seen that the bridge and the railway can be detected and the boudaries of the river can be identified well.

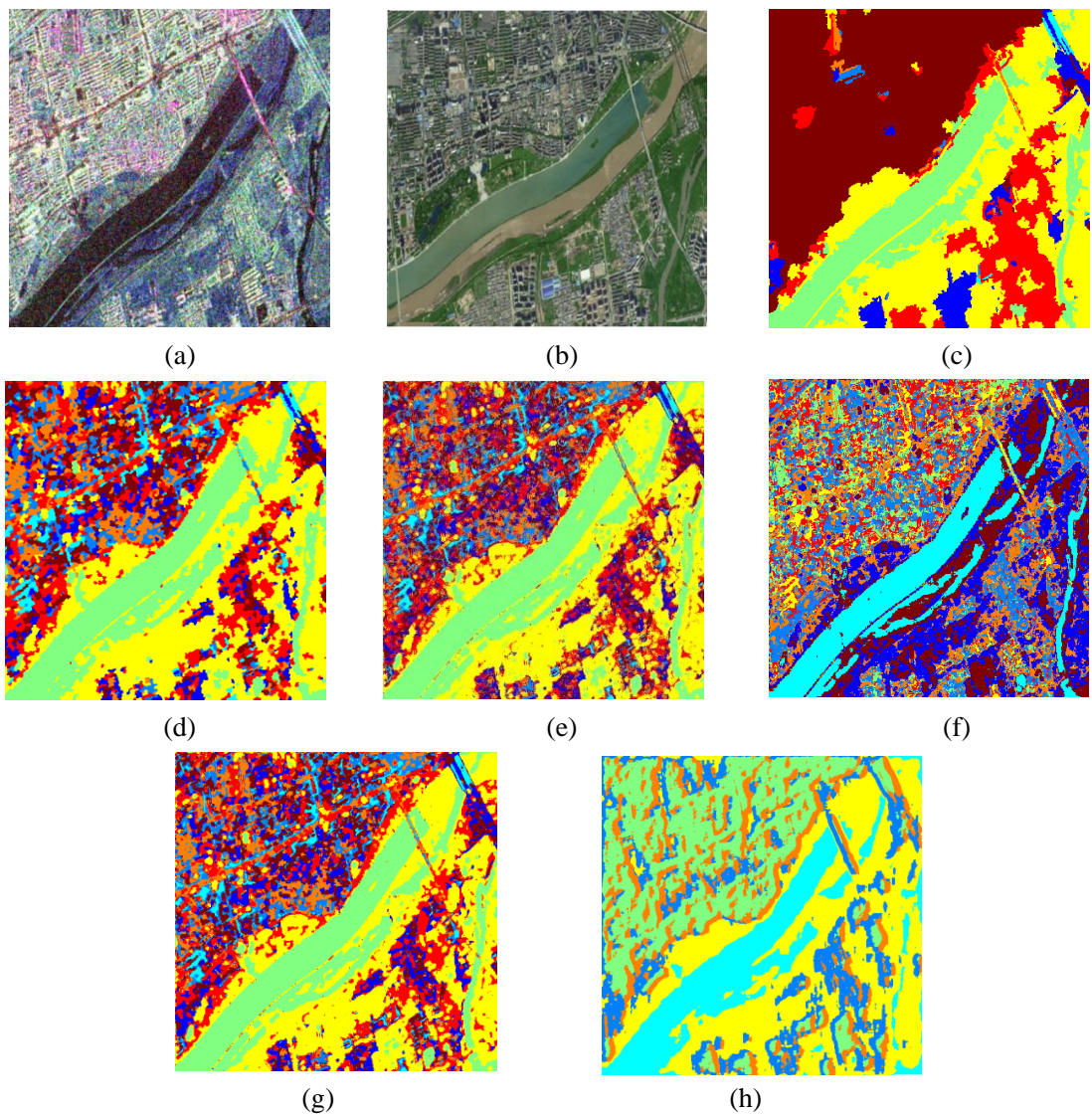

Fig. 16. Classification results of Xi'an area. (a) Pseudo color image of Xi'an area with Pauli base. (b) Satellite image from Google Earth. (c) Classification by PHS_SM method. (d) Classification by MSSM method. (e) Classification by H/α-Wishart. (f) Classification by SPECR method. (g) Classification by Wishart based MRF method. (h) Deep learning classification result.

## F. Parameter Analysis

In our method, $K$ is a parameter which affects the classification accuracy. Fig. 17 indicates the effect of $K$ on classification accuracy in the E-SAR data shown in Fig. 14(a). $K$ varies from 3 to 21 with the increment of 2 by experiments. It can be seen that the correct rate is low with small $K$ and keep flat later, and the maximum value is derived when $K$=9. It means the PHS_SM method is robust to $K$ when $K$ is not too small or too large. That's because the aggregated degree cannot be estimated correctly with too small or too large $K$. Therefore, we select $K$=9 for Fig. 14(a).

Moreover, the region number $N_r$ is set by the user and it affects the classification accuracy. The effect of $N_r$ on the classification accuracy is illustrated in Fig. 18. The E-SAR data in Fig. 14(a) is used for test. The total region number of homogenous areas is 81. $N_r$ varies from 10 to 50 with the increment of 5. Fig. 18 shows that the classification accuracy rate is low with a small $N_r$ and almost keeps flat after 30. It demonstrates that the over-segmentation can produce better performance than under-segmentation. To guarantee an over-segmented result, we set $N_r$ as a slightly large value.

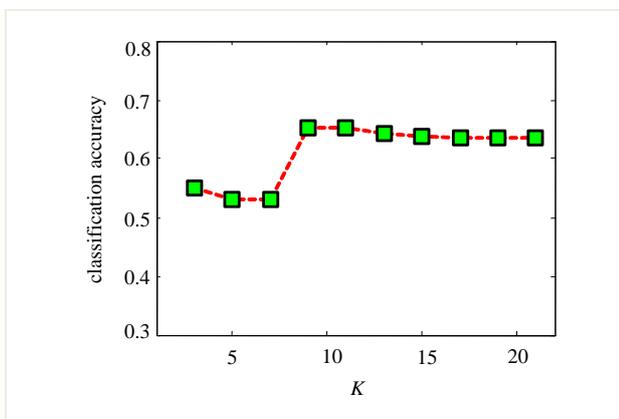 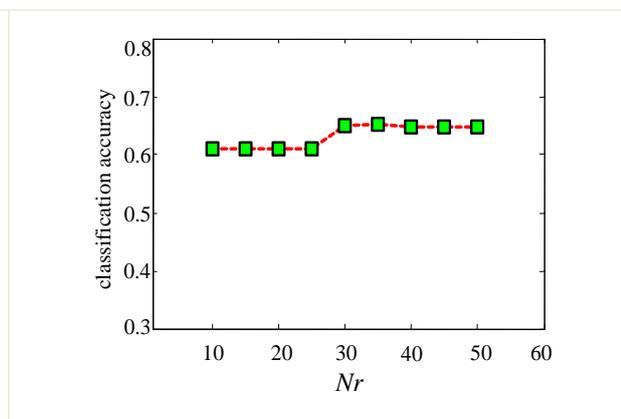

Fig. 17. Effect of *K* on classification accuracy.   Fig. 18. Effect of $N_r$ on classification accuracy.

## VI CONCLUSION

In this paper, a novel hierarchical semantic model has been proposed for POLSAR terrain classification. To learn the structure-level representation of vision, a primal-level semantic is constructed by a polarimetric sketch map which is composed of sketch segments. Moreover, to partly simulate the concept-level representation of vision, a middle-level semantic is formulated by a region map which partitions an image into aggregated, structural and homogenous regions. In addition, different schemes are designed for three regions by considering their characteristics. Furthermore, a semantic-polarimetric classification method takes advantages of both the semantic information and the scattering mechanism of PolSAR images to obtain more accurate terrain classification. Experimental results have demonstrated that our method gets better performance than the existing methods in region homogeneity and boundary preservation especially for complex terrain types.

In addition, the region map has a wider range of application such as the PolSAR image segmentation, speckle reduction, object detection of PolSAR images and so forth. Our method provides a new framework for PolSAR image processing by partitioning a complex scene into three relatively homogenous subspaces and designing different models for three subspaces.

However, in the proposed method, there are some parameters which should be selected by the user. Some adaptive division methods with fewer parameters will be exploited in the future work. Besides, more spatial relationship of sketch segments can be exploited for PolSAR images in the future.

# APPENDIX

## 1. Symbol Definition

Table I   symbols in the algorithm of aggregated region extraction.

| symbol | definition |
|---|---|
| $U = \bigcup_{k=1}^{m} T_k$ | aggregated segment set |
| $m$ | number of the aggregated segment subset |
| $T_k = \{s_1, s_2, \cdots, s_{n_k}\}$ | an aggregated segment subset |
| $|T_k| = n_k$ | segment number in $T_k$ |
| $T = \{T_1, T_2, \cdots T_k, \cdots\}$ | grown aggregated segment subsets |
| $\delta_2$ | spatial constraint threshold |
| $r_i$ | aggregated region from $T_i$. |
| $R = \{r_1, r_2, \cdots, r_m\}$ | total aggregated regions |
| $\emptyset$ | empty set |

## 2. Procedure of Aggregated Region Extraction

| Procedure of aggregated region extraction algorithm |
|---|
| Input: AS set U. |
| Output: aggregated regions R= {$r_1, r_2,..., r_m$} |
| Initialization: T=∅, $T_i$ = ∅. |
| |
| While $U - T \neq \varnothing$ |
|    Select $\varphi_i$ from $U - T$ as a seed, $T_i$ = {$\varphi_i$}. |
|    While $\varphi_i$ in $T_i$ is not traversed |
|       For j=1:k |
|          Select the *j*th nearest neighbor $\varphi_j$ of $\varphi_i$ |
|            If $d_{ij} <= \delta_2$ & $\varphi_j \notin T_i$ ($d_{ij}$ is the distance between $\varphi_i$ and $\varphi_j$) |
|              Add $\varphi_j$ to $T_i$ |
|            End |
|       End |
|       $\varphi_i$ is labeled as traversed |
|    End |
|    If $|T_i| < K$ |
|       Label segments in $T_i$ as IS |
|    else |
|       i=i+1; |
|       Closing operator is used to $T_i$ and an aggregated region $r_i$ is formed.   %Stage 2 |
| End |

# ACKNOWLEDGEMENT

This work was carried out with the part-supports of the National Basic Research Program (973 Program) of China (No.2013CB329402), the National Natural Science Foundation of China (No. 61173090), the Fund for Foreign Scholars in University Research and Teaching Programs (the 111 Project) (No. B07048), the National Science Foundation of China under Grant (Grant No. 91438103 and 91438201), the Program for Cheung Kong Scholars and Innovative Research Team in University (No.IRT1170), the Fundamental Research Funds for the Central Universities (No. JB140317).